\documentclass[journal]{IEEEtran}

\usepackage{listings}

\usepackage{graphicx}
\usepackage{multirow} 
\usepackage{array}
\usepackage{overpic}

\usepackage{hyperref}
\hypersetup{hypertex=true,
            colorlinks=true,
            linkcolor=blue,
            anchorcolor=blue,
            citecolor=blue}
\usepackage{tcolorbox}

\usepackage{amsmath} 
\usepackage{amssymb}
\usepackage[T1]{fontenc}

\begin{document}

\title{Robot Detection System 3: LRF groups and Coordinate System}

\author{Jinwei~Lin,~\IEEEmembership{Member,~IEEE, ACM}
        
\thanks{Jinwei Lin is come from Monash University,} 
\thanks{ORCID of Jinwei Lin:0000 0003 0558 6699,} 
\thanks{Manuscript written in 2018.}} 

\markboth{Journal or Conference of \LaTeX\, No.~1, May~2023}
{Shell \MakeLowercase{\textit{et al.}}: Simple  Arrow Area Architecture Template}
\maketitle


\begin{abstract}
Front-following is more technically difficult to implement than the other two human following technologies, but front-following technology is more practical and can be applied in more areas to solve more practical problems. In this paper, we will analyze the detailed design of LRF groups, the structure and combination design of coordinate system of Robot Detection System. We use enough beautiful figures to display our novel design idea. Our research result is open source in 2018, and this paper is just to expand the research result propagation granularity. Abundant magic design idea are included in this paper, more idea and analyzing can sear and see other paper naming with a start of Robot Design System with Jinwei Lin, the only author of this series papers.
\end{abstract}




\section{Introduction}
The LRF group is a new design concept that we have established. In the study, we set the horizontal ground where the robot is located to the XY plane, and the direction perpendicular to the horizontal ground to the Z-axis direction. The LRF group is a combination of two LRF detectors in the same Z-axis direction. The two LRF detectors in the LRF group work independently and do not interfere with each other. The two LRF detectors of the same LRF group are in the same straight line when viewed in the Z-axis direction.

\section{Structural Design}
\begin{figure}[t]
\centering
\includegraphics[width=0.99\columnwidth]{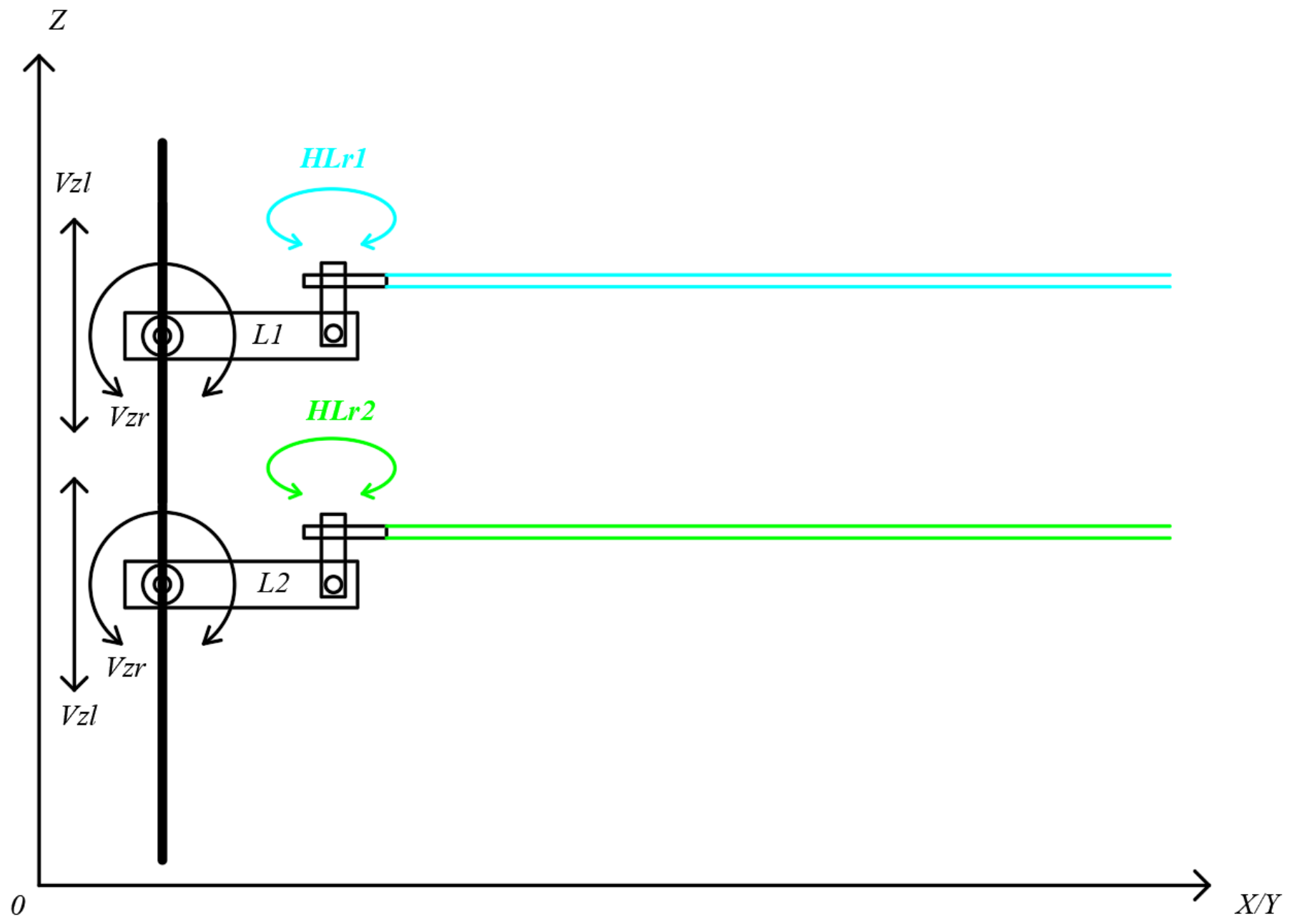}
\caption{Schematic diagram of the structure and operation of the LRF group.}
\label{fig12}
\end{figure}

As shown in Figure\ref{fig12}, two LRF sensors are distributed in a straight line parallel to the Z axis. In general, the hardware performance of these two LRFs is not required to be the same. Therefore, the robot designer can install different performance LRF sensors according to actual needs. In general, the LRFs in the LRF group satisfies the freedom of motion action as long as there are two degrees of freedom. One degree of freedom is used to rotate in a plane parallel to the Z axis. Vzr in Figure 1 represents a rotating electric machine capable of precise angle control, which corresponds to this degree of freedom. Another degree of freedom is the degree of freedom of the LRF to make a rotational motion parallel to the plane of the support bar. This degree of freedom can control the rotating electric machine to perform precise angular rotation. The support bar refers to a rotatable crossbar that supports the LRF, such as L1 and L2 in Figure\ref{fig12}. As shown in Figure\ref{fig12}, if we look in the direction parallel to the XY plane, the lights emitted by the two LRFs in the LRF group seen is in a horizontal straight line, but they are actually a plane. These two planes become the laser ray plane. This plane is made up of a number of planar areas made from the laser lights emitted from the LRFs.

\section{Detection Mode}
Why do we want to introduce the new concept and design idea of the LRF group? This is closely related to the detection mode of our robot's sensor system. We defined two scan modes for the LRF group. One is the normal scan mode and the other is the locking scan mode. The following algorithm and design analysis will be performed on a robot based on a sensor system with a four-sided central model.

\subsection{Data Storage Mode}
After scanning with the detector, the computer system of the robot needs to store and process the acquired scan data. Here we define two storage processing methods for the robot's computer system, one is obscured scanning storage, and the other is map planning storage. If the computer system performs obscured scan storage, the scanned data will not be permanently recorded, and the scan data of the scan area other than the edge of the largest scan area will be recorded when the distance from the scan area is large to a certain extent. It will be cleared by the system. The purpose of this is to save the memory resources of the computer system of the robot and reduce the burden of data processing of the computer system. Forgotten storage is suitable for robots with uncertain working areas. Such robots often work in uncertain surroundings, so the storage of surrounding environment information does not help the robot's following actions and cruise movements. The robot system in the map planning storage mode records all the data in the scanning area within the farthest edge distance that the LRF can detect, and generates a corresponding area map by the map generation algorithm. Map-planned storage is suitable for fixed or relatively fixed robots at work. Such robots often move in the same area. So the more familiar the area is, the more the robot has sufficient data to make a more reasonable judgment.

\subsection{Normal Mode}
\begin{figure}[t]
\centering
\includegraphics[width=0.99\columnwidth]{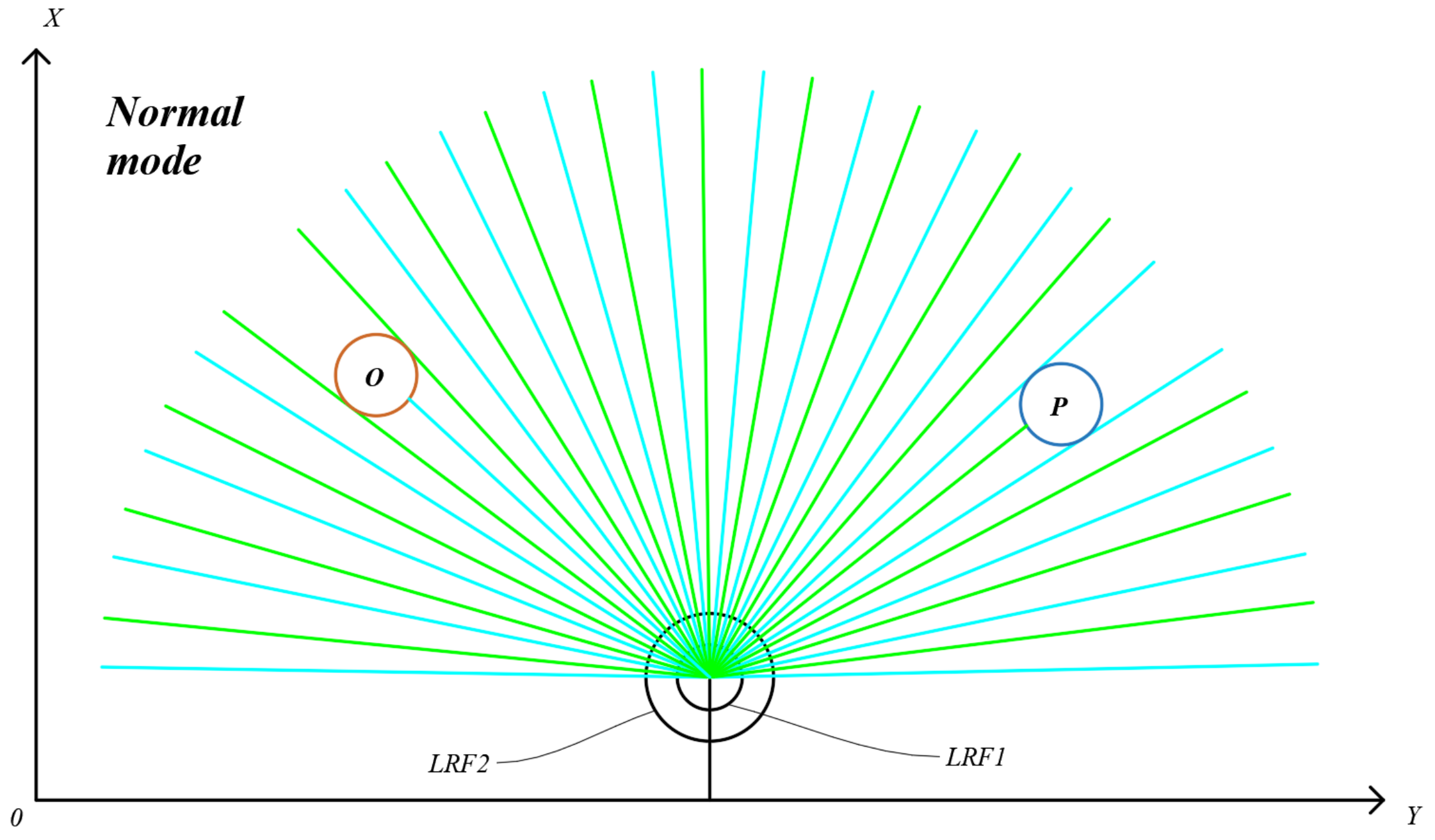}
\caption{Normal scan mode for the LRF group.}
\label{fig13}
\end{figure}

In the normal scan mode, the two LRFs in the LRF group are in the same working mode, that is, a 180-degree angle scan. When the target person is not in the detection area corresponding to the LRF group, the sensor system will automatically start the normal scan mode. As shown in Figure\ref{fig13}. The representative obstacle marked "O", marked as "P" represents other humans except the target person, called irrelevant human. The two LRFs of the LRF group are represented by LRF1 and LRF2, respectively. The green and sky blue rays in the figure represent the detected rays emitted by LRF1 and LRF2, respectively. Obstacle or unrelated humans may be mobile or stationary. In order to save computing resources, it is possible to perform obscured scan storage only for obstacles or unrelated humans. If the robot is working in a specific area, the robot will perform map planning storage.

In the normal mode, the two LRFs of the same LRF group are in the same working mode, so the data processing method of the double scan data superposition detection can be used to reduce the detection errors of the LRFs. The basic idea of double-scan data overlay detection is to detect the same position in the detection area by using two LRFs at different positions, and then perform average error processing on the scan results. The easiest way to handle the average errors is to find the arithmetic means of the coordinates.

\subsection{Locking Mode}
The second scan mode is the locking scan mode. In this mode, once the robot finds that the target person enters the detection area corresponding to a certain LRF group, the robot's scanning processing mechanism automatically sets the LRF group corresponding to the scanning area to enter the locking scan mode. In this mode, the algorithm mechanism will assign an LRF in the LRF group for a locked tracking scan of the target person. At the same time, the algorithmic mechanism will allocate another LRF in the LRF group for scanning the obstacles and unrelated humans. As shown in FIGURE 14, the LRFs in the locking scan mode scan only in the area corresponding to the range of the detection angle at which the target person is located and the corresponding nearby area. The basic design principle of the lock-type scanning is to perform round-trip angle scans on the scanning area including the target person and the vicinity thereof through the LRF, thereby acquiring the detection scan data information of the edge of the target person in real time. In the process of locked scanning, if the target person moves, his or her motion data information will be recorded in high speed and high precision, and stored in the corresponding database to form tracking path data information with a certain format. This data information will be used to predict the future paths of the target person. In general, normal scanning does not record the detected data information. Once the robot in the locked scan mode detects that the target person has moved in position, the scan angle range of the locked scan will be adjusted to the corresponding angle area in time. In Figure 14, the target person is represented by a circle with an "H" mark. When the target person moves from point H to point H', the corresponding scanning angle of the target person is also changed from 10 degrees to 18 degrees. The detection and acquisition data system controls the LRF in the locked scan mode in real time to perform corresponding angle adjustment and scanning to ensure that the target person is always in the locked scan angle range during the whole scanning process. 

At the same time, the path information of the target person will be collected and recorded in real time. We use a straight line parallel to the Y-axis as the starting line, counterclockwise as the rotation direction, and rotate until the first boundary line of the scanning area. The angle of rotation is recorded as the deflection angle corresponding to the scanning area. Then, $\theta1$ is the deflection angle when the target person is at the "H" point, and $\theta2$  is the deflection angle when the target person is at the "H" point. Since the scanning angle is only the angle at which the first boundary line scanned to the object is recorded, the shape and position information of the object cannot be completely reflected. Therefore, in order to better describe and solve the problem of scanning angle range, we define the following concepts:

1. The first boundary angle: the angle from the reference polar axis to the angle at which the detected light is tangent to the edge of the object being detected for the first time. It is also the angle between the detected light and the reference polar axis when the detected light first contacts the detected object, and is denoted as $\Psi_a$;

2. The angle of the second boundary: the angle from the reference pole axis to the angle at which the detected light is tangent to the edge of the object being detected. It is also the angle between the detected light and the reference polar axis when the last time the detected light touches the detected object, which is denoted as $\Psi_b$;

3. Scan angle range: the absolute value of the difference between the angle between the first boundary and the second boundary;

4. Scanning angle interval: an angle interval formed by the angle between the first boundary angle and the second boundary, including an angle between the first boundary angle and the second boundary. Generally, the first boundary angle is the starting point.

\begin{figure}[t]
\centering
\includegraphics[width=0.99\columnwidth]{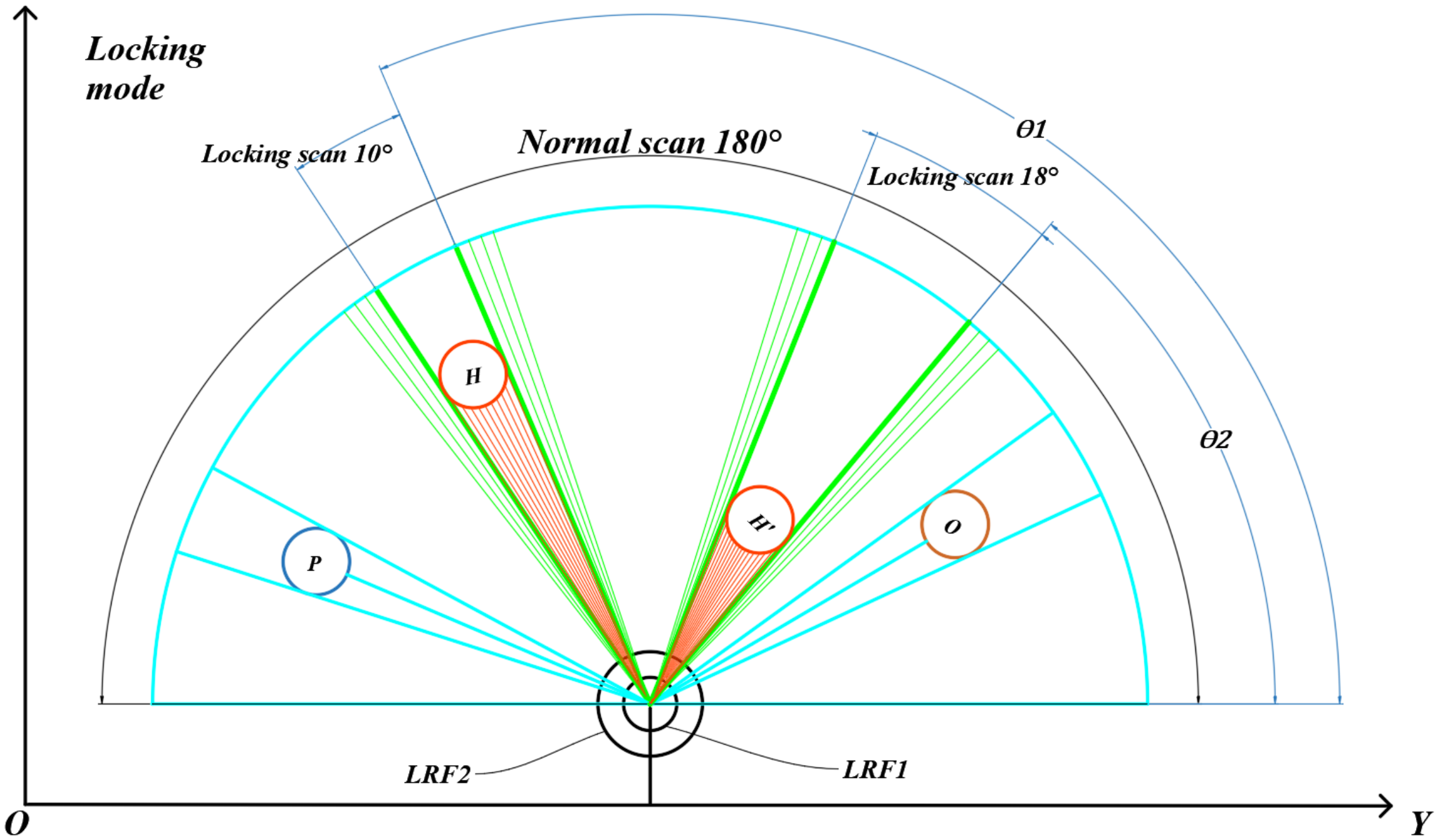}
\caption{Schematic diagram of the structure and operation of the LRF group.}
\label{fig14}
\end{figure}

In a normal round-trip rotation scan, there are generally only two boundary lines, the start boundary line and the end boundary line. Correspondingly, we call them the first dividing line and the second dividing line respectively. The width of the width of the corresponding detected portion of the detected object in a certain scanning direction can be determined by the two boundary lines. As shown in Figure\ref{fig14}, the scanning angle interval corresponding to the target person H is ($\Theta1$, $\Theta1$+10), and the scanning angle interval corresponding to the target person H' is ($\Theta2$, $\Theta2$+18). N.B.: The angle between the first boundary angle and the second boundary is the important parameters in the detection algorithm.

\section{Combination Coordinate System}
After the scan mode is determined, the next step will be to scan the detected object with the determined scan model and obtain the corresponding scan data of the detected object. To accurately scan the object being inspected and obtain the corresponding highly accurate scan data information, it is necessary to establish a scientific and rigorous coordinate system. Next, we will introduce a combined coordinate system suitable for the robot of this project. The combined coordinate system is constructed by combining Cartesian coordinate system and spherical coordinate system.

After the scan mode is determined, the next step will be to scan the detected object with the determined scan model and obtain the corresponding scan data of the detected object. To accurately scan the object being detected and obtain accurate scan data information, it is necessary to establish a scientific and rigorous coordinate system. Next, we will introduce a combined coordinate system suitable for the robot of this project. The combined coordinate system is constructed by combining Cartesian coordinate system and spherical coordinate system.

\begin{figure}[t]
\centering
\includegraphics[width=0.99\columnwidth]{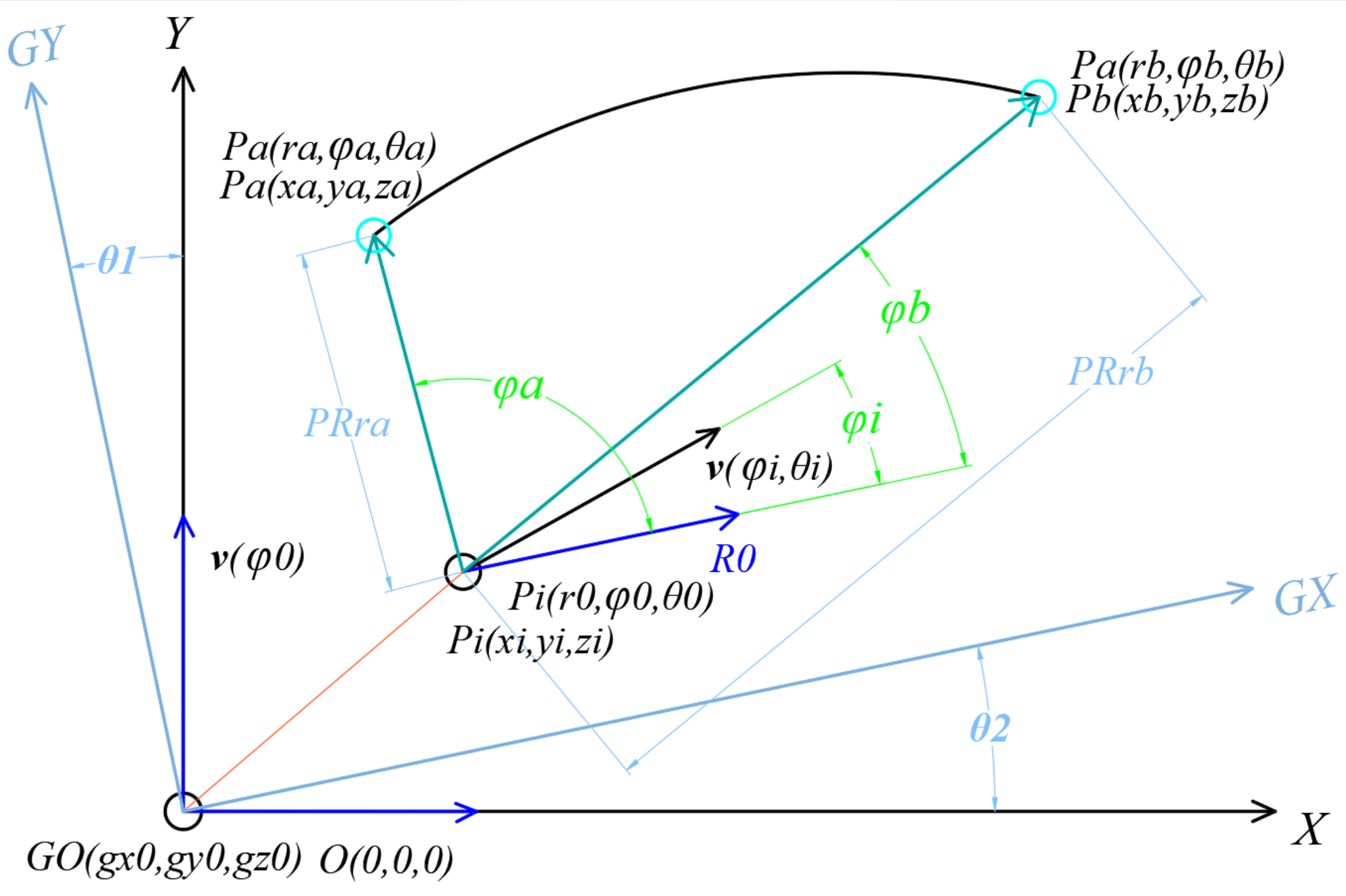}
\caption{Design schematic of a combined coordinate system for a scanned data model.}
\label{fig15}
\end{figure}

As shown in Figure\ref{fig15}, the pale blue GXGYGZ is a global positioning coordinate system. That is to say, GXGYGZ is determined by geomagnetic field, GPS and other global navigation systems or indoor positioning and navigation. The coordinate system does not change with the position and orientation of the robot. For GXGYGZ, which is a global positioning coordinate system, we set GX's direction to be positive north, GY's 30direction to be positive east, and GZ's direction to be vertical sea level. The origin O of GXGYGZ global positioning coordinate system is the geometric center origin when the robot starts to depart after accepting the task. V(0) is the forward orientation of the robot when it starts to depart after accepting the task. Once the robot starts to perform a new follow task, the forward orientation of the robot will be set to the Y axis, and the right orientation of the robot will be set to the X axis. Similarly, the starting point of the coordinate system is the geometric center point O of the robot, so as to construct the scanning and positioning coordinate system XYZ of the robot. The origin of the XYZ scanning positioning coordinate system is set to (0, 0, 0). The origin coordinates of GXGYGZ coordinate system are set as (gx, gy, gz), where GX is the latitude value of the corresponding position points when the robot starts the task, GY is the longitude value of the corresponding position points when the robot starts the task, and GZ is the sea level height value of the corresponding position points when the robot starts the task. It can be seen that the coordinate axis Z of the XYZ coordinate system coincides with the coordinate axis GZ of the GXGYGZ coordinate system, but the coordinate axis (X, Y) of the XYZ coordinate system does not necessarily coincide with the coordinate axis (GX, GY) of the GXGYGZ coordinate system.

We define the angle between the GX axis and the X axis as the counterclockwise plane angle between the GXGYGZ coordinate system and the XYZ coordinate system, expressed in $\Theta$2. The angle between the GY axis and the Y axis is also defined as the counterclockwise plane angle between the coordinate system GXGYGZ and the coordinate system XYZ, which is expressed in $\Theta$1. That is $\Theta$2= $\Theta$1. Assuming that the robot is in the Pi point in space during the process of following the target person, the coordinates of the Pi point are expressed as Pi (xi, yi, zi) in the XYZ coordinate system. R0 is the reference axis parallel to the GX axis, and the direction is the same as that of GX. In the process of scanning, Pi (xi, yi, zi) is used as the origin point Pi(r0, $\Phi$0, $\Theta$0) of the robot in the process of performing the following task, and the spherical coordinate system is established. R0 is the projection of the polar axis of the spherical coordinate system on the ZY plane, that is, the polar axis itself. Note that R0 always keeps its direction parallel to the GX axis of GXGYGZ, pointing at the same direction, while the robot follows the target person. v($\Phi$i,$\Theta$i) is the velocity vector of the robot at Pi point. PRra and PRrb respectively detect the projection of laser ray RA and Rb on the XY plane. The anticlockwise angles of PRra and PRrb and polar axis R0 are a and b, respectively. The spherical coordinates of Pa (xa, ya, za) and Pb (xb, yb, zb) in the XYZ coordinate system are Pa (ra, $\Phi$a, $\Theta$a) and Pb (rb, $\Phi$b, $\Theta$b), respectively.

Next we will discuss the conversion between the GXGYGZ positioning coordinate system, the XYZ following coordinate system and the R$\Phi$$\Theta$ scanning spherical coordinate system.

Set the starting point of the robot as the starting point. The GXGYGZ coordinate corresponding to the origin of origin is (gx0, gy0, gz0), and $\theta$1 = $\theta$2 = $\theta$g measured by the angle sensor. For a point Pi(xi, yi, zi) in the XYZ coordinate system, the coordinates of the point PGi corresponding to the GXGYGZ coordinate system are shown as Equation\ref{eq7}.

\begin{equation}
\label{eq7}
\begin{cases}gx=\sqrt{xi^2+yi^2}\times\cos(arccos\left(\frac{xi}{\sqrt{xi^2+yi^2}}\right)-\theta g)+gx0\\gy=\sqrt{xi^2+yi^2}\times\sin(arccos\left(\frac{xi}{\sqrt{xi^2+yi^2}}\right)-\theta g)+gy0\\gz=zi+gz0&\end{cases}
\end{equation}

For the point PGi (gx, gy, gz) on the GXGYGZ coordinate system, to convert to a point Pi(xi, yi, zi) in the XYZ coordinate system, the corresponding coordinates are shown as Equation\ref{eq8}.

\begin{equation}
\label{eq8}
\begin{cases}
xi= \frac{\sqrt{(gx-gx0)^2+(gy-gy0)^2}}{\cos(arcsin\left(\frac{gx}{\sqrt{gx^2+gy^2}}\right)+\theta g)}\\
yi=\frac{\sqrt{(gx-gx0)^2+(gy-gy0)^2}}{\sin(arcsin\left(\frac{gx}{\sqrt{gx^2+gy^2}}\right)+\theta g)}\\
zi=gz-gz0
\end{cases}
\end{equation}

Set the robot's geometric center as the origin when the robot first starts performing the following task. The corresponding coordinates of the origin in the XYZ coordinate system are (0, 0, 0). In the process of the robot following the target character, the coordinate point of the XYZ following coordinate system corresponding to a certain LRF in a certain LRF of the robot is Pi(xi, yi, zi), and some of the XYZ coordinate system space detected by the LRF The XYZ coordinate of a probe point is Pa(xa, ya, za). Taking the Pi point as the origin, the Pi point is parallel to the reference polar axis R0 of the GX axis, and the angle between R0 and the X axis is $\theta$g. It can be seen that the angle between R0 and the X axis is the angle between the GX axis and the X axis (i.e., $\theta$g). Taking Pi as the origin and R0 as the reference polar axis, the R$\Phi$$\theta$ detection spherical coordinate system is established. That is to say, the origin of the R$\Phi$$\theta$ detection spherical coordinate system is not set by the geometric center of the robot. (IE: Therefore, there is a three-dimensional spatial translation process when transforming the coordinate system. For the specific processing and related algorithms, please refer to the analysis part of the mutual conversion between the coordinate systems of the combined coordinate system.) At this time, the laser is detected by the sensor. The zenith angle corresponding to the beam irradiation point Pa is $\theta$a, the azimuth angle is $\Phi$a, and the radial distance is ra (ie, the length of the laser ranging beam). N.B.: In the combined coordinate system, the GXGYGZ positioning coordinate system and the XYZ following coordinate system are determined after setting the starting point of the robot to the origin of the respective coordinate system. The origin of the coordinate system will not change in the entire follow-up task of the robot. The origin of the R$\Phi$$\theta$ detection spherical coordinate system (r0=0, $\Phi$0=0, $\theta$0=0) changes as the position of the LRF of the robot changes. I.E.: Changes with the XYZ following coordinates of the LRF of the LRF group of the robot or the change of the GXGYGZ positioning coordinates. That is to say, the R$\Phi$$\theta$ detection spherical coordinate system is dynamically changed to have real-time performance.

Because the GXGYGZ positioning coordinate system and the origin of the XYZ following coordinate system are set relative to the geometric center of the robot when considering the precise origin position setting, and the origin of the R$\Phi$$\Theta$ detecting spherical coordinate system is set being relative to an exit point of the detected light of an LRF in the LRF group of robot. Therefore, if the R$\Phi$$\Theta$ detection spherical coordinate system wants to convert the corresponding coordinates with the other two coordinate systems, the origin translation operation must be performed on the R$\Phi$$\Theta$ detection spherical coordinate system. It is assumed that the XYZ coordinates corresponding to the light exit point of the LRF used to establish the origin of the R$\Phi$$\Theta$ detecting spherical coordinate system are (Xl, Yl, Zl). The R$\Phi$$\Theta$ coordinate corresponding to a certain detection point detected by the R$\Phi$$\Theta$ detection spherical coordinate system with the light exit point of the LRF as the origin is (rt, $\Phi$t, $\Theta$t). The detection point is converted into the coordinates of the XYZ coordinate system with the light exit point of the LRF as the origin (xlt, ylt, zlt), and the detected point is converted into the coordinate of the XYZ coordinate system with the geometric center point of the robot as the origin (x, y, z), then have the following relationship as shown in Equation\ref{eq10}.

\begin{equation}
\label{eq10}
\begin{cases}x=xlt-XL\\y=ylt-YL\\z=zlt-ZL&\end{cases}
\end{equation}

For a point Pa(xa, ya, za) in the XYZ coordinate system, on the R$\Phi$$\Theta$ spherical coordinate system with the point Pi(xi, yi, zi) as the origin, the point corresponding to the Pa point is the point PSa. (ra, $\Phi$a, $\Theta$a), the coordinates of the point PSa are shown as Equation\ref{eq11}.

\begin{equation}
\label{eq11}
\begin{cases}ra=\\
\sqrt{(xa-XL-xi)^2+(ya-YL-yi)^2+(za-ZL-zi)^2}\\\varphi a=\varphi a\\\theta a=\theta a&\end{cases}
\end{equation}

Then, for the point PSa(ra, $\Phi$a, $\Theta$a) on the R$\Phi$$\Theta$ spherical coordinate system with the point Pi(xi, yi, zi) as the origin, in the XYZ coordinate system with the geometric center of the robot as the origin, the coordinate point corresponding to point PSa is Pa(xa, ya, za), and the coordinates of point Pa are as shown as Equation\ref{eq12}.

\begin{equation}
\label{eq12}
\begin{cases}
rx=ra\times sin\theta a\times cos\varphi a\\ry=ra\times sin\theta a\times sin\varphi a\\rz=ra\times cos\theta a\\xa=rx\div cos\theta g-ry\times sin\theta g+xi+XL\\ya=ry\div cos\theta g+rx\times sin\theta g+yi+YL\\za=rz+zi+ZL
\end{cases}
\end{equation}

In the following principles and algorithm descriptions in this report, unless otherwise specified, both of the “XYZ coordinate system” or “XYZ following coordinate system” refer to the XYZ Cartesian coordinate system with the geometric center of the robot as the origin.

The combined coordinate system we built consists of three coordinate combinations and is built in three dimensions. We propose to use the coordinate system in three-dimensional space instead of the two-dimensional space, mainly because, from the current technological development, our robots live in the three-dimensional world and work in the three-dimensional world. Therefore, the three-dimensional coordinate system is more suitable for the robot detection system. Of course, building a three-dimensional coordinate system is more complicated than constructing a two-dimensional coordinate system, but the three-dimensional coordinate system will be more realistic, can handle more problems, and acquire more detection data for more accurate and complex detection and control, as well as provide more complete and accurate data and information for the control or predictive models.

Another noteworthy question is how the sensor system acquires the value of the Z axis. The Z value in the GXGYGZ positioning coordinate system is directly obtained from the height sensor detection in the robot body. The Z value of the XYZ coordinate system is algorithmically converted from the value detected by the height sensor in the robot body. Below we will focus on why we should introduce R$\Phi$$\Theta$ to detect the spherical coordinate system instead of the cylindrical coordinate system or other coordinate system when dealing with this problem. 

If the analysis is from static and dynamic angles, it can be considered that the R$Phi$$\Theta$ detection spherical coordinate system is dynamic because its origin position coordinates are changed every moment during the execution of the robot following the character. The GXGYGZ positioning coordinate system and the XYZ following coordinate system are static, because in the process of the robot performing the following character, its origin position coordinate is unchanged after the start time is determined.

\begin{figure}[t]
\centering
\includegraphics[width=0.99\columnwidth]{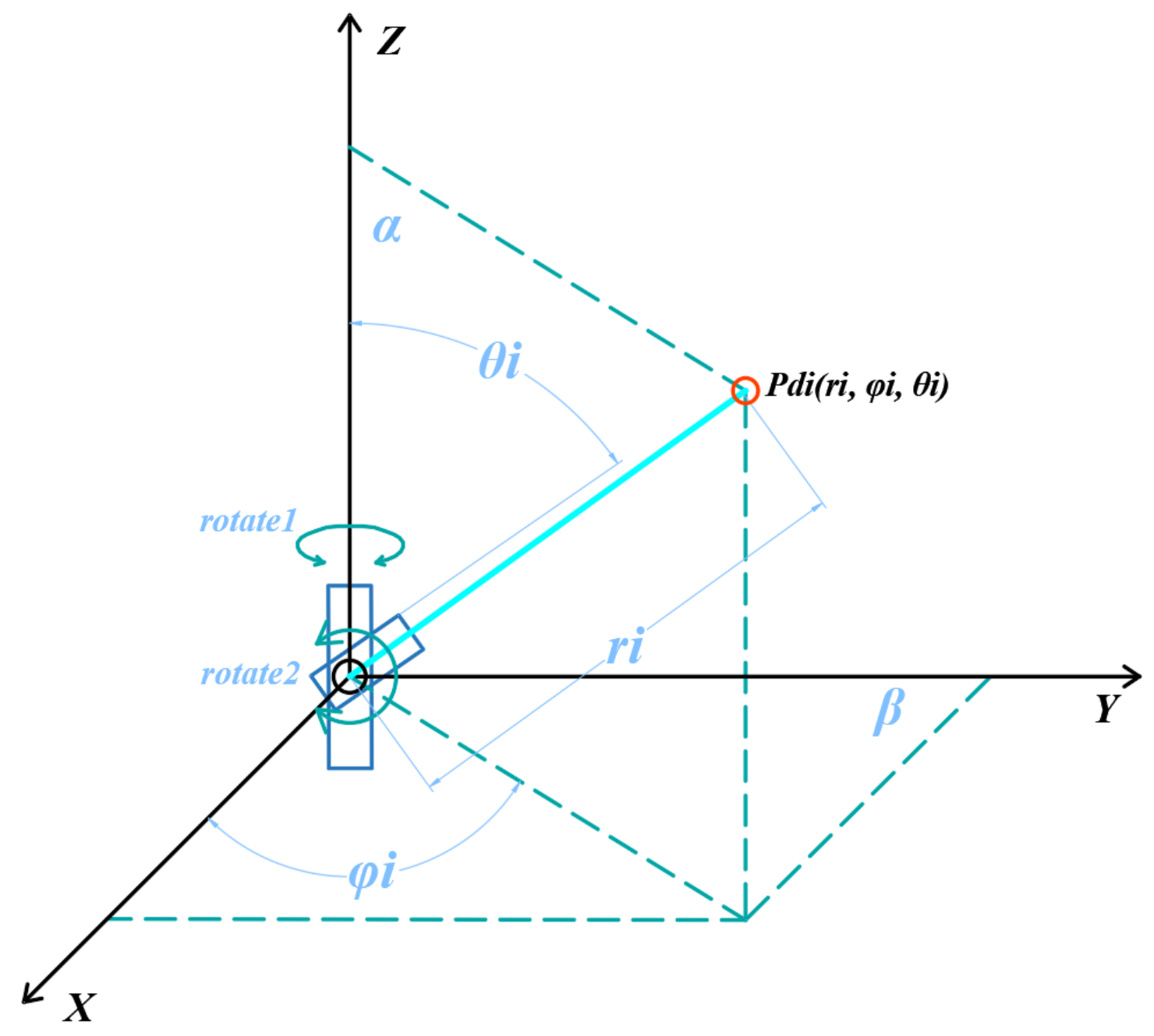}
\caption{Relationship between R$\Phi$$\Theta$ detection spherical coordinate system and LRF detection group.}
\label{fig16}
\end{figure}

The R$\Phi$$\Theta$ detection spherical coordinate system is the most important component of the combined coordinate system. When n detection, we introduce R$\Phi$$\Theta$ to detect the spherical coordinate system instead of other coordinate systems. The important purpose is to closely combine with the operating mechanism and structure of the sensor system we designed. We believe that the introduction of R$\Phi$$\Theta$ to detect the spherical coordinate system can make the detection system of the robot of this project reach its maximum potential. In this project, each LRF in the LRF group in the robot sensor system has at least two degrees of freedom. In general, only two degrees of freedom are required. One degree of freedom is used for rotational motion in a plane parallel to the Z-axis, and the other degree of freedom is used to rotate within the cylinder centered on the Z-axis, the points of rotation being located in a series of planar clusters parallel to the XOY plane . As shown in Figure\ref{fig16}, it is assumed that the point Pdi$(ri, \phi i, \theta i)$ is a point detected by the LRF, and rotate1 and rotate2 respectively indicate the rotation modes corresponding to the two degrees of freedom. The origin of the XYZ coordinate system is set to LRF. At the same time, LRF is also the origin of the R$\Phi$$\Theta$ detection spherical coordinate system, and is also the emission point of the LRF detection light. By the LRF's rotary control unit, we can easily get the values of $\phi$i and $theta$i by rotating the angle sensor of the unit, and it is directly obtained. And because the original function of the LRF is to measure the point to-point linear distance, and ri represents the length of the detected light emitted by the LRF, this measurement distance can be directly obtained. Therefore, three important parameters of ri, $\phi$i and $theta$i can be obtained directly. This greatly improves the convenience and accuracy of the test data. This is why we want to introduce R$\Phi$$\Theta$ to detect the spherical coordinate system for the final detection.

\begin{figure}[t]
\centering
\includegraphics[width=0.99\columnwidth]{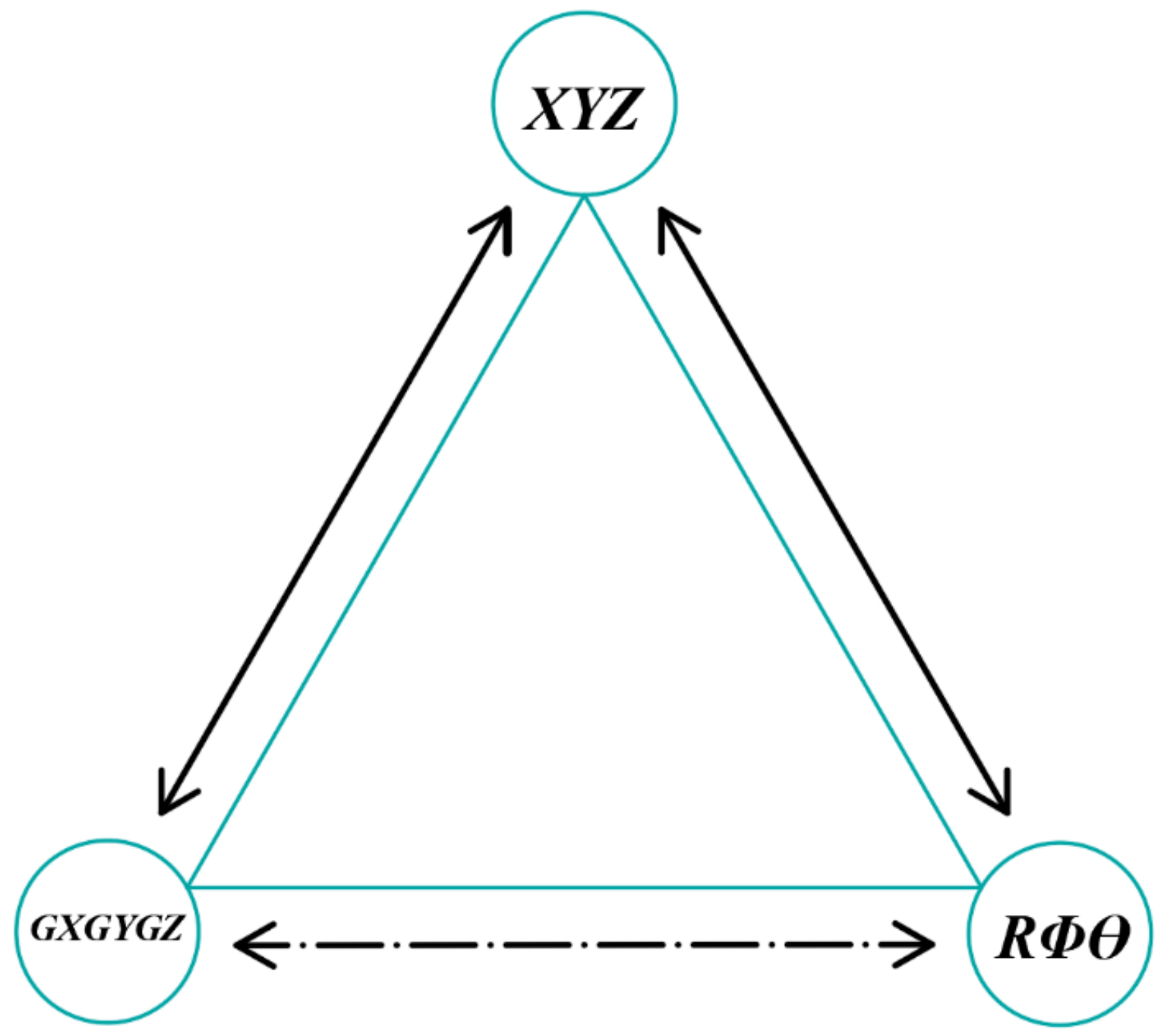}
\caption{Mutual conversion diagram of each coordinate system of the combined coordinate system.}
\label{fig17}
\end{figure}

After we knowing how to realize transforming the coordinates of a point in the GXGYGZ positioning coordinate system to the coordinates of a corresponding point in the XYZ following coordinate system, and how to realize transforming the corresponding coordinates of a point in the XYZ following coordinate system to the coordinates of a corresponding point in R$\Phi$$\Theta$ detecting the spherical coordinate system, we can establish a bridge of the transformation of the coordinates of the two corresponding points between the two coordinate systems. As shown in Figure\ref{fig17}, the XYZ following coordinate system will function as a link bridge in the coordinate transformation of the three coordinate systems. The solid line in Figure 17 indicates that direct conversion is possible, and the broken line indicates indirect conversion. If the coordinate points in the GXGYGZ positioning coordinate system are to be converted into the corresponding coordinate points in the R$\Phi$$\Theta$ detection spherical coordinate system, they can be first converted into the corresponding coordinate points in the XYZ following coordinate system, and then converted into the corresponding coordinate point in the R$\Phi$$\Theta$ detection spherical coordinate system. Similarly, if the coordinate points in the R$\Phi$$\Theta$ detection spherical coordinate system are to be converted into the corresponding coordinate points in the GXGYGZ positioning coordinate system, they can be first converted into the corresponding coordinate points in the XYZ following coordinate system, and then converted into the corresponding coordinate point in the GXGYGZ positioning coordinate system.

\begin{figure}[t]
\centering
\includegraphics[width=0.99\columnwidth]{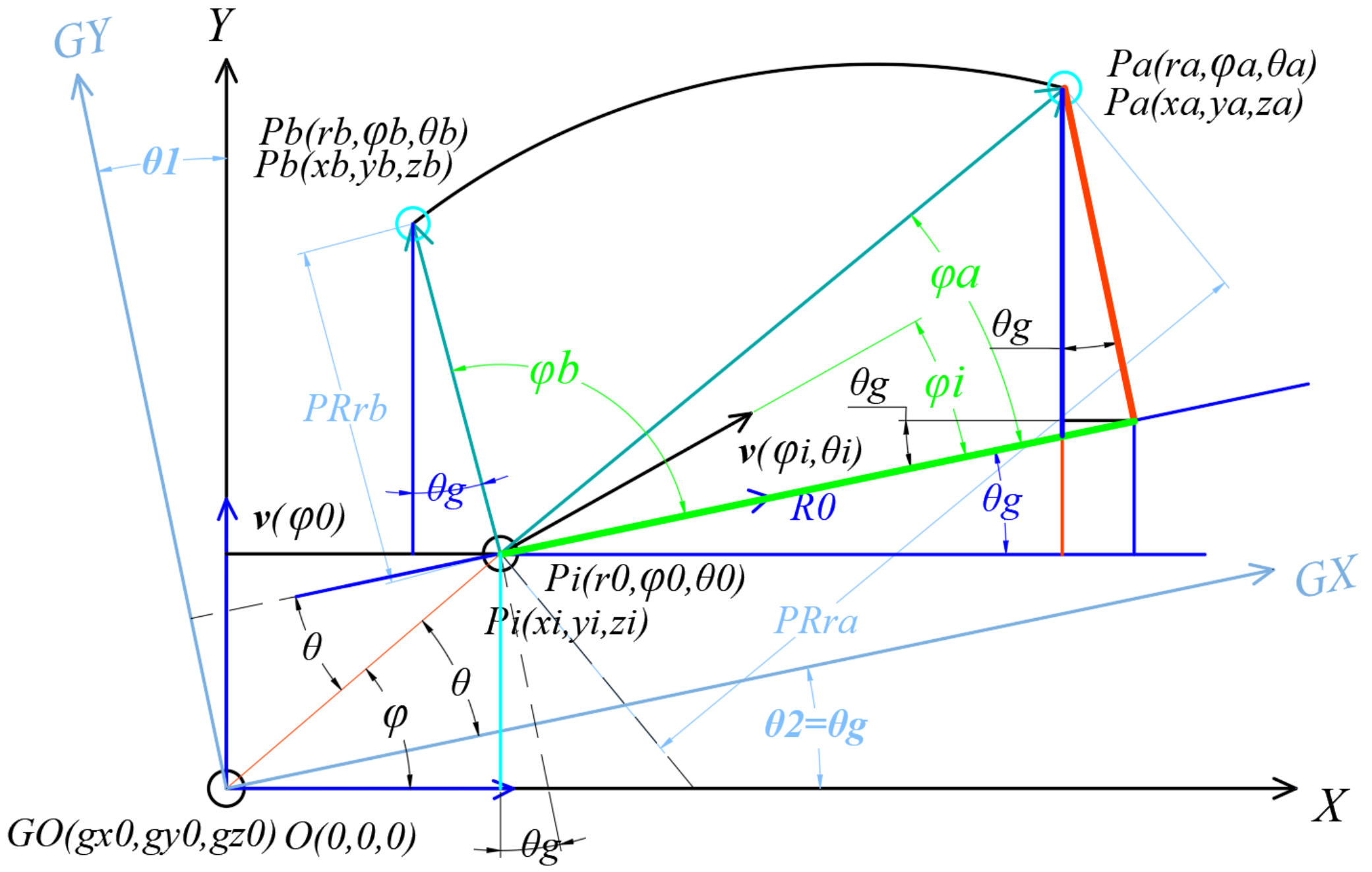}
\caption{Schematic diagram of the derivation process of the combined coordinate system.}
\label{fig18}
\end{figure}

The derivation process in which the coordinates of the corresponding points in the combined coordinate system are mutually converted is not the focus of this project report, so it is omitted in this version of the report. A schematic diagram of the derivation calculation is given here, as shown in Figure\ref{fig18}, for the reader's reference.

\section{Conclusion}
For content arrangement of this paper series, we make the first simple conclusion in this secession. In this paper, we made detailed analysis and discussion of the LRF groups definition and made a detailed mathematical principle analysis for the coordinate system design of this robot detection. For more detailed technologies analyzes, note the other papers of this paper series. Sensor system design is significant for robot detection system design, for example, combine with the area drawing detection technology \cite{lin2022multi}, can improve the detection function of robot detection system. For 3D human movement pose generation \cite{10072686}, robot detection system with the methods above will be useful in some situations.

\bibliographystyle{IEEEtran}
\bibliography{ref}{}
\end{document}